\documentclass[wcp]{jmlr}

\usepackage{longtable}% for long tables

\usepackage{graphicx}
\usepackage{microtype}
\usepackage{graphicx}
\usepackage{booktabs}
\usepackage{amsfonts}
\usepackage{amsmath}
\usepackage{pgfplots}
\usepackage{multirow}

\pgfplotsset{compat=1.14}

\DeclareMathOperator*{\argmin}{arg\,min}
\DeclareMathOperator*{\argmax}{arg\,max}
\newcommand\numberthis{\addtocounter{equation}{1}\tag{\theequation}}

\jmlrvolume{101}
\jmlryear{2019}
\jmlrworkshop{ACML 2019}

\title[Stochastic Gradient Trees]{Stochastic Gradient Trees}

\author{%
  \Name{Henry Gouk} \Email{hgouk@inf.ed.ac.uk}\\
  \addr School of Informatics \\ University of Edinburgh \\ Edinburgh, United Kingdom
  \AND
  \Name{Bernhard Pfahringer} \Email{bernhard.pfahringer@waikato.ac.nz}\\
  \Name{Eibe Frank} \Email{eibe.frank@waikato.ac.nz}\\
  \addr Department of Computer Science \\ University of Waikato \\ Hamilton, New Zealand
 }

\editors{Wee Sun Lee and Taiji Suzuki}

\begin{document}

\maketitle

\begin{abstract}
We present an algorithm for learning decision trees using stochastic gradient information as the source of supervision. In contrast to previous approaches to gradient-based tree learning, our method operates in the incremental learning setting rather than the batch learning setting, and does not make use of soft splits or require the construction of a new tree for every update. We demonstrate how one can apply these decision trees to different problems by changing only the loss function, using classification, regression, and multi-instance learning as example applications. In the experimental evaluation, our method performs similarly to standard incremental classification trees, outperforms state of the art incremental regression trees, and achieves comparable performance with batch multi-instance learning methods.
\end{abstract}

\section{Introduction}
% Introduce problem we are solving and motivation for it
%   Online gradient-based optimisation has allowed deep learning and linear models to applied to a huge number of different problems with very little effort
%       Show some examples
%   This work addresses the issue of incrementally learning decision trees using gradient-based optimisation, thus enabling their application to a wide variety of tasks.
Stochastic gradient descent is the workhorse of contemporary machine learning. Methods for scalable gradient-based optimisation have allowed deep neural networks to tackle a broad range of problems, from binary classification to playing video games~\citep{lecun2015}. The scalable nature of incremental methods like stochastic gradient descent enable training on very large datasets that cannot fit in memory, and combining it with automatic differentiation allows one to solve new tasks by simply changing the loss function. In contrast, applying other model classes, such as decision trees, to new tasks requires the design of a new optimisation algorithm that can search for models that perform well on the new problem. Moreover, if one intends to train such a model on a large dataset, this optimisation algorithm must scale well. As such, designing a general purpose algorithm for incrementally constructing decision trees that minimise arbitrary differentiable loss functions would be of great interest to the machine learning community. Such an algorithm would enable decision trees to be applied to a broad range of problems with minimal effort.

% Discuss the drawbacks of existing solutions
%   Hoeffding trees must be changed significantly for each new task
%       Must change split heuristic
%       Must prove bound on split heuristic in order to use Hoeffding bound
%   Gradient boosting techniques have the correct interface
%       They construct an ensemble, resulting in very large model sizes
%       Not incremental, so they require a lot of compute power to train on large datasets
%   We propose a solution that avoids these problems
Hoeffding trees~\citep{domingos2000} are one approach for incrementally constructing decision trees, but they lack the generality of the method proposed in this paper. In order to adapt the Hoeffding tree induction algorithm to new tasks other than classification, one must select a new heuristic for measure the quality of splits, and also prove an upper bound on this measure in order to apply the Hoeffding inequality. Conversely, the gradient boosting literature has provided many examples of how one can construct an ensemble of decision trees using arbitrary differentiable loss functions~\citep{friedman2001, chen2016, ke2017}. However, the resulting models are typically very large---often containing hundreds, or sometimes thousands of trees. Constructing these ensembles generally requires significant computing resources and highly optimised implementations, even for modestly sized datasets~\citep{mitchell2017}. In this paper we propose stochastic gradient trees (SGT), which is both general and scalable. The generality comes from the ability to optimise for arbitrary differentiable loss functions, and the scalability is due to the ability of this algorithm to incrementally build a single tree using gradient information, rather than constructing a large ensemble.

% Overview of the tasks we apply SGTs to (and why they are worth solving!)
%   Streaming classification and regression
%   Multi-instance learning
%   Overview of results
Several tasks are used to demonstrate the broad applicability of SGTs. Firstly, it is demonstrated that SGTs perform similarly to Hoeffding trees when applied in the streaming classification setting---where the learner may only see each instance once during training. Following this, we compare SGTs trained with the squared error loss function to several variants of Hoeffding trees that are specialised for regression. These experiments show that SGTs achieve state of the art performance on challenging streaming regression problems. Lastly, multi-instance learning (MIL) is considered. In this problem, one is supplied with ``bags'' during training. Each bag contains several feature vectors, but only a single label. If a bag is labelled positive, then at least one instance inside the bag is a positive example. Otherwise, all training examples in the bag are negative. In this setting, SGTs exhibit competitive performance to specialised batch multi-instance learning methods.

% Outline contributions very clearly
%   Incremental gradient-based decision tree induction
%   Use of t-test instead of Hoeffding bound, which removes requirement of bounded range on heuristic
%   Demonstrate how it can be applied to classification, regression, multi-instance (weakly supervised?) learning
To summarise, our contributions are three-fold: (i) we show how incremental decision trees can be adapted to use stochastic estimates of the gradient as the source of supervision. (ii) To remove the requirement of deriving bounds on the gradients and Hessians of each new loss function, we demonstrate how $t$-tests can be used in place of the Hoeffding inequality when splitting a node. (iii) It is demonstrated how our novel incremental decision tree can be applied to streaming classification, regression, and multi-instance learning.

\section{Related Work}
% Examples of repurposing Hoeffding trees for different tasks
Hoeffding trees~\citep{domingos2000} are a commonly used technique for incremental learning. In each leaf node, they maintain a co-occurrence histogram between feature values and classes in order to determine the quality of potential splits through the use of a standard split quality measure, such as the information gain. The Hoeffding concentration inequality is applied to determine whether there is enough evidence to identify the best split, or if more training examples must been before a split can be perfored. Various modifications for Hoeffding trees exist that enable them to solve problems other than classification. The FIMT-DD~\citep{ikonomovska2011a} and ORTO~\citep{ikonomovska2011} are Hoeffding tree variants designed for streaming regression problems. FIMT-DD makes use of linear models in the leaf node of each tree to increase the precision of predictions, while ORTO utilises option nodes to enable each instance to travel down multiple paths in the tree. \citet{read2012} and \citet{mastelini2019} extend Hoeffding trees to address streaming  multi-label classification and multi-target regression problems, respectively. For each of these modification, a new measure of split quality must be chosen, and an upper bound derived so that the Hoeffding inequality can be applied. Our method automates both of these steps by leveraging gradient information from arbitrary loss functions to measure split quality, and using a different hypothesis test to determine if a node is ready to be split.

% Previous work on learning trees from gradients
Previous work on learning decision trees with differentiable loss functions has focused on the use of soft splits. \citet{suarez1999} first use a conventional decision tree induction method (the CART method of \citet{breiman1984}) to find the structure of the tree. Following this, each hard split is converted into a soft split by replacing the hard threshold with a logistic regression model. These soft splits are subsequently fine-tuned using a process similar to backpropagation. While this continuous approximation of discrete models improves performance on regression tasks, it does not enable one to train models on new problems, as the structure of the trees are still learnt using a traditional tree induction technique. More recently, \citet{yang2018} focus on learning interpretable models by training quasi-soft decision trees using backpropagation, and subsequently converting soft splits into hard splits. The depth of the tree is determined by the number of features in the dataset, and the differentiable method for learning tree structure requires storing a proper $n$-ary tree. This, coupled with a learned discretization method, causes the algorithm to scale very poorly with the number of features in the dataset. In contrast to these methods, our approach is able to efficiently learn the structure of tree models, as well as select the appropriate features and thresholds at each split.

Gradient boosting~\citep{friedman2001} is a technique that can be used to construct an ensemble of classifiers trained to minimise an arbitrary differentiable loss function. Each addition of a new model to the ensemble can be thought of as performing a Newton step in function space. There are two main differences between the method presented in this paper and methods such as XGBoost~\citep{chen2016} and LightGBM~\citep{ke2017}. Firstly, our method learns incrementally, whereas these popular gradient boosting approaches are inherently batch learning algorithms. Secondly, SGTs are not an ensemble technique: when constructing a stochastic gradient tree a Newton step is performed with every update to a single tree, rather than each new tree in the ensemble. Updates to an SGT take the form of splits to leaf nodes or updates the prediction value of a leaf node.

\section{Stochastic Gradient Trees}
In supervised incremental learning, data is of the form $(\vec x_t, y_t) \in \mathcal{X} \times \mathcal{Y}$, a new pair arrives at every time step, $t$, and the aim is to predict the value of $y_t$ given $\vec x_t$. Algorithms for this setting must be incremental and enable prediction at any time step---they cannot wait until all instances have arrived and then train a model. In this section, we describe our method for incrementally constructing decision trees that can be trained to optimise arbitrary twice-differentiable loss functions. The first key ingredient is a technique for evaluating splits and computing leaf node predictions using only gradient information. Secondly, to enable loss functions that have unbounded gradients, we employ standard one-sample $t$-tests rather than hypothesis tests based on the Hoeffding inequality to determine whether enough evidence has been observed to justify splitting a node.

\subsection{Leveraging Gradient Information}
\label{sec:gradient-info}
We assume a loss function, $l(y, \hat{y})$, that measures how well our predictions, $\hat{y}$, match the true values, $y$. Predictions are generated using an SGT, optionally composed with an activation function, $\sigma$,
\begin{equation}
    \hat{y} = \sigma(f(\vec x)).
\end{equation}
Training should minimize the expected value, as estimated from the data observed between the current time step, $t$, and time step, $r$, at which the tree was previously updated. Assuming i.i.d. data, the expectation can be stochastically approximated using the most recent observations,
\begin{equation}
\mathbb{E}\lbrack l(y, \hat{y}) \rbrack \approx \frac{1}{t - r}\sum_{i=r+1}^t l(y_i, \hat{y_i}).
\end{equation}
The predictions, $\hat{y_i}$, are obtained from the SGT, $f_t$. At each time step, we aim to find a modification, $u:\mathcal{X} \to \mathbb{R}$, to the tree that takes a step towards minimising the expected loss. Because $f_t$ is a decision tree, $u$ will be a function that represents a possible split to one of its leaf nodes, or an update to the prediction made by a leaf: the addition of $f_t$ and $u$ is the act of splitting a node in $f_t$, or changing the value predicted by an existing leaf node. Formally, the process for considering updates to the tree at each time step is given by
\begin{equation}
f_{t+1} = f_t + \argmin_{u} \lbrack \mathcal{L}_t(u) + \Omega(u) \rbrack,
\end{equation}
\noindent where
\begin{equation}
\mathcal{L}_t(u) = \sum_{i=r+1}^t l(y_i, f_t(\vec x_i) + u(\vec x_i)),
\end{equation}
\noindent and
\begin{equation}
	\Omega(u) = \gamma |Q_u| + \frac{\lambda}{2} \sum_{j \in Q_u} v_u^2(j).
\end{equation}

The $\Omega$ term is a regularizer, $Q_u \subset \mathbb{N}$ is the set of unique identifiers for the new leaf nodes associated with $u$, and $v_u:\mathbb{N} \to \mathbb{R}$ maps these new leaf node identifiers to the difference between their predictions and the prediction made by their parent. The first term in $\Omega$ imposes a cost for each new node added to the tree, and the second term can be interpreted as a prior that encourages the leaf prediction values to be small. In our experiments, we set $\lambda$ to $0.1$ and $\gamma$ to $1$. In the case of Hoeffding trees, and also SGTs, only the leaf that contains $\vec x_t$ will be considered for splitting at time $t$, and information from all previous instances that have arrived in that leaf will be used to determine the quality of potential splits. The algorithm also has the option to leave the tree unmodified if there is insufficient evidence to determine the best split.

There are two obstacles to incrementally training a tree using an arbitrary loss function. Firstly, the splitting criterion must be designed to be consistent with the loss to be minimised. Secondly, the leaf prediction values of the leaf nodes must be chosen in a manner that is consistent with the loss. Both problems can be overcome by adapting a trick used in gradient boosting techniques~\citep{friedman2001, chen2016, ke2017} that expands an ensemble of trees by applying a Taylor expansion of the loss function around the current state of the ensemble. We only consider modification of a single tree, therefore the empirical expectation of the loss function can be approximated using a Taylor expansion around the unmodified tree at time $t$:
\begin{equation}
\mathcal{L}_t(u) \approx \sum_{i=r+1}^t \lbrack l(y_i, f_t(\vec x_i)) + g_i u(\vec x_i) + \frac{1}{2}h_i u^2(\vec x_i) \rbrack,
\end{equation}
\noindent where $g_i$ and $h_i$ are the first and second derivatives, respectively, of $l$ with respect to $f_t(\vec x_i)$. Optimisation can be further simplified by eliminating the constant first term inside the summation, resulting in
\begin{align*}
\Delta \mathcal{L}_t(u) &= \sum_{i=r+1}^t \lbrack g_i u(\vec x_i) + \frac{1}{2}h_i u^2(\vec x_i) \rbrack \numberthis \\
						&= \sum_{i=r+1}^t \Delta l_i(u),
\end{align*}
\noindent which now describes the change in loss due to the split, $u$.

This function is evaluated for each possible split to find the one that yields the maximum reduction in loss. As in the Hoeffding tree algorithm, at time $t$, we only attempt to split the leaf node into which $\vec x_t$ falls, and we consider splitting on each attribute. For each potential split, we need to decide what values should be assigned to any newly created leaf nodes. Note that we also consider the option of not performing a split at all, and only updating the prediction made by the existing leaf node.

We introduce some notation to explain our procedure. Firstly, we define what a potential split looks like:
\begin{equation}
u(\vec x) =
\begin{cases}
v_u(q_u(\vec x)), & \text{if} \,\, \vec x \in \text{Domain}(q_u) \\
0, & \text{otherwise}
\end{cases}
\end{equation}
\noindent where $q_u$ maps an instance in the current leaf node to an identifier for a leaf node that would be created if the split were performed. We denote the codomain of $q_u$---the set of identifiers for leaf nodes that would created as a result of performing this split---as $Q_u$. We define the set $I_u^j$ as the set of indices of the instances that would reach the new leaf node identified by $j$. The objective can then be rewritten as
\begin{equation}
\Delta \mathcal{L}_t(u) = \sum_{j \in Q_u} \sum_{i \in I_u^j} \lbrack g_i v_u(j) + \frac{1}{2}h_i v_u^2(j) \rbrack,
\end{equation}
\noindent which can be rearranged to
\begin{equation}
\label{eq:split-objective}
\Delta \mathcal{L}_t(u) = \sum_{j \in Q_u} \lbrack (\sum_{i \in I_u^j} g_i) v_u(j) + \frac{1}{2}(\sum_{i \in I_u^j} h_i) v_u^2(j) \rbrack,
\end{equation}
which uses the sums of the gradient and Hessian values that have been seen thus far. The optimal $v_u(j)$ for each candidate leaf can be found by taking the relevant term in Equation~\ref{eq:split-objective} and adding the corresponding term from $\Omega$,
\begin{equation}
	(\sum_{i \in I_u^j} g_i) v_u(j) + \frac{1}{2}(\sum_{i \in I_u^j} h_i) v_u^2(j) + \frac{\lambda}{2} v_u^2(j),
\end{equation}
\noindent then setting the derivative to zero,
\begin{equation}
	0 = (\sum_{i \in I_u^j} g_i) + (\lambda + \sum_{i \in I_u^j} h_i) v_u(j),
\end{equation}
\noindent and solving for $v_u(j)$, yielding
\begin{equation}
v_u^\ast(j) = -\frac{\sum_{i \in I_u^j} g_i}{\lambda + \sum_{i \in I_u^j} h_i}.
\end{equation}

Viewing the expected loss as a functional, this induction procedure can be thought of as performing Newton's method in function space. In gradient boosting, the addition of each new tree to the ensemble performs a Newton step in function space. The difference in our approach is that each Newton step consists of modifying a prediction value or performing a single split, rather than constructing an entire tree. For loss functionals that cannot be perfectly represented with the quadratic approximation in Newton-type methods, an SGT can potentially take advantage of gradient information more effectively than trees trained using conventional gradient tree boosting~\citep{chen2016,ke2017}.

\subsection{Splitting on Numeric Attributes}
When splitting on nominal attributes, we create a branch for each value of the attribute, yielding a multi-way split. We deal with numeric attributes by discretizing them using simple equal width binning. A sample of instances from the incoming data is used to estimate the minimum and maximum values of each numeric attribute---if these are not already known in advance. Any future values that do not lie in the estimated range are clipped. The number of bins and the number of instances used to estimate the range of attribute values are user-provided hyperparameters. In our experiments, we set them to 64 and 1,000, respectively. Given a discretized attribute, we consider all possible binary splits that can be made based on the bin boundaries, thus treating it as ordinal~\citep{frank1999}.

\subsection{Determining when to Split}
Equation~\ref{eq:split-objective} estimates the quality of a split but does not indicate whether a split should be made. Hoeffding trees use the Hoeffding concentration inequality to make this decision. It states that, with some probability $1 - \delta$,
\begin{equation}
\mathbb{E}\lbrack \overline{X} \rbrack > \overline{X} - \epsilon,
\end{equation}
\noindent with
\begin{equation}
\epsilon = \sqrt{\frac{R^2 \text{ln}(1/\delta)}{2n}},
\end{equation}
\noindent where $\overline{X}$ is the sample mean of a sequence of random variables $X_i$, $R$ is the range of values each $X_i$ can take, and $n$ is the sample size used to calculate $\overline{X}$. Suppose the best split considered at time $t$ is $u^a$. Let $\overline{L} = \frac{1}{n}\hat{\mathcal{L}_t}(u^a)$ be the mean change in loss if the split were applied, as measured on a sample of $n \leq t$ instances. Thus, if $-\overline{L} > \epsilon$, we know, with $1 - \delta$ confidence, that applying this split will result in a reduction of loss on future instances.

In order to apply the Hoeffding bound, we must know the range, $R$, of values that can be taken by the $n$ terms, $\Delta \hat{l_i}$, in $\Delta \hat{\mathcal{L}}$. In our application, this would require proving upper and lower bounds of the first and second derivatives of the loss function, and constraining the output of the tree to lie within some prespecified range, thus preventing rapid experimentation with different loss functions for novel tasks---one of the properties of deep learning that enables such a diverse set of tasks to be solved. To circumvent this problem, we instead use Student's $t$-test to determine whether a split should be made. The $t$ statistic is computed by
\begin{equation}
t = \frac{\overline{L} - \mathbb{E}\lbrack \overline{L} \rbrack}{s / \sqrt{n}},
\end{equation}
\noindent where $s$ is the sample standard deviation of $L_i$ and, under the null hypothesis, $\mathbb{E}\lbrack \overline{L} \rbrack$ is assumed to be zero---i.e., it is assumed that the split does not result in a change in loss. A $p$ value can be computed using the inverse cumulative distribution function of the $t$ distribution and, if $p$ is less than $\delta$, the split can be applied.

This test assumes that $\overline{L}$ follows a normal distribution. Although it cannot be assumed that each $L_i$ will be normally distributed, due to the central limit theorem, we are justified in assuming $\overline{L}$ will be normally distributed for sufficiently large $n$. Computing $s$ requires estimating the sample variance of $L_i$, which is made easier by initially considering each of the new leaf nodes, $j \in Q_u$, in isolation:
\begin{equation}
	\label{eq:loss-variance}
	\text{Var}(L_i) = \text{Var}(G_i v_u(j) + \frac{1}{2} H_i v_u^2(j)),
\end{equation}
\noindent where $G_i$ and $H_i$ are the random variables representing the gradient and Hessian values, respectively. We intentionally treat $v_u(j)$ as a constant, even though this ignores the correlation between the prediction update values and the gradient and Hessian values. Empirically, this does not appear to matter, and it eliminates the need to compute the variance of a quotient of random variables---an expression for which there is no distribution-free solution.

Equation~\ref{eq:loss-variance} cannot be computed incrementally because the $v_u(j)$ are not known until all the data has been seen. It is also infeasible to store all of the gradient and Hessian pairs because this could lead to unbounded memory usage. Instead, the equation can be rearranged using some fundamental properties of variances to yield
\begin{equation}
	\text{Var}(L_i) = v_u^2\text{Var} (G_i) + \frac{1}{4} v_u^4 \text{Var}(H_i) + v_u^3 \text{Cov}(G_i, H_i),
\end{equation}
\noindent where we have dropped the ``$(j)$'' for compactness. The variances and covariances associated with each feature value can be incrementally estimated using the Welford's method~\citep{welford1962} and one of the algorithms presented by~\citet{bennett2009}, respectively. When considering splits, the sample statistics associated with each feature value---and therefore each new leaf node---can be aggregated using the concurrent estimation algorithms given by~\citet{bennett2009}, resulting in the sample variance, $s^2$.

The process used to determine whether enough evidence has been collected to justify a split would be prohibitively expensive to carry out every time a new instance arrives. In practice, we follow the common trend in online decision tree induction and only check whether enough evidence exists to perform a split when the number of instances that have fallen into a leaf node is a multiple of some user specified parameter. As with many incremental decision tree induction implementations, this value is set to 200 by default.

\section{Example Tasks}
This section outlines the tasks and associated loss functions used to demonstrate the generality of stochastic gradient trees in the experimental evaluation. For completeness, the first and second derivatives of each loss function are given, though we note that it would be very easy to incorporate an automatic differentiation system into an SGT implementation to remove the need to derive these manually.

\subsection{Streaming Classification}
Streaming classification is a variant of the typical classification problem encountered in machine learning where a learning algorithm is presented with a continuous stream of examples, and must efficiently update the model with the new knowledge obtained at each time step. In this setting, the length of the stream is unknown, and predictions can be requested from the model at any time. This means the model must be trained incrementally. In this paper, the multiclass streaming classification problem is addressed using a committee of SGTs, where one tree is trained for each class. This committee is composed with a softmax function, so the probability that an instance, $\vec x_i$, belongs to class $j$ is estimated by
\begin{equation}
    \hat{y}_{i,j} = \frac{\text{exp}\{f_j(\vec x_i)\}}{\sum_{c=1}^{k} \text{exp}\{f_c(\vec x_i)\}},
\end{equation}
where $f_c$ is the SGT trained to predict a real-valued score for class $c$, and $k$ is the number of classes. In practice, we hard-wire $f_k(\vec x) = 0$ in order to reduce the number of trees being trained. The categorical cross entropy loss function is used to train this model,
\begin{equation}
    \ell^{CE}(\vec y, \hat{\vec y}) = -\sum_{c=1}^{k} y_{c} \text{log} (\hat{y}_j),
\end{equation}
where $\vec y$ is the ground truth label encoded as a one-hot vector. The first derivative is,
\begin{equation}
    \frac{\delta \ell^{CE}}{\delta f_c(\vec x)} = \hat{y}_c - y_c,
\end{equation}
and the second derivative is given by
\begin{equation}
    \frac{\delta^2 \ell^{CE}}{\delta f_c^2(\vec x)} = \hat{y}_c (1 - \hat{y}_c).
\end{equation}

\subsection{Streaming Regression}
Streaming regression is very similar to streaming classification, but a numeric value must be predicted instead of a nominal value. A single SGT, $f$, is used to generate predictions, $\hat{y}$, and trained using the squared error loss function,
\begin{equation}
    \ell^{SE}(y, \hat{y}) = \frac{1}{2}(\hat{y} - y)^2,
\end{equation}
which has first derivative
\begin{equation}
    \frac{\delta \ell^{SE}}{\delta f(\vec x)} = \hat{y} - y,
\end{equation}
and second derivative
\begin{equation}
    \frac{\delta^2 \ell^{SE}}{\delta f^2(\vec x)} = 1.
\end{equation}

\subsection{Multi-Instance Learning}
Multi-instance learning is a specific instantiation of weakly-supervised learning where a bag of training instances is assigned a single binary annotation. Under the standard MIL assumption (see~\citet{foulds2010a} for more details), a positive label indicates that the bag contains at least one instance belonging to the positive class, while a negative label means that the bag does not contain any instances from the positive class. The ability to train instance-level classifiers from bag-level supervision can significantly reduce the resources required to annotate a training set, and this particular setting maps well to several tasks in computer vision and computational chemistry. For example, learning to classify whether an image subwindow contains an object category of interest using only image-level labels, rather than bounding boxes, is an active area of research in the computer vision community. In this case, the image is the bag, and each subwindow in the image is an instance.

Suppose $X_i$ is a bag of instances and $f$ is an SGT, the probability that $X_i$ contains a positive instance is estimated by
\begin{equation}
    \hat{y}_i = \frac{1}{1 + \text{exp}(-\max_{\vec x \in X_i} f(\vec x))}.
\end{equation}
The binary cross entropy loss function is used to optimise the model,
\begin{equation}
    \ell^{BCE}(y, \hat{y}) = -y \text{log}(\hat{y}) - (1-y) \text{log} (1-\hat{y}),
\end{equation}
where $y$ is the ground truth label for the bag. The first and second derivatives of this loss with respect to $f(\vec x)$ for each $\vec x \in X_i$ are
\begin{equation}
    \frac{\delta \ell^{BCE}}{\delta f(\vec x)} =
    \begin{cases}
        (\hat{y}_i - y_i) & \text{if} \quad \vec x = \argmax_{z \in X_i} f(\vec z) \\
        0 & \text{otherwise}
    \end{cases}
\end{equation}
and
\begin{equation}
    \frac{\delta^2 \ell^{BCE}}{\delta f^2(\vec x)} =
    \begin{cases}
        \hat{y}_i (1 - \hat{y}_i) & \text{if} \quad \vec x = \argmax_{z \in X_i} f(\vec z) \\
        0 & \text{otherwise}.
    \end{cases}
\end{equation}

In contrast to the previous example tasks, which operate in the data stream mining setting, the multi-instance learning tasks considered in the experimental evaluation are batch learning problems. As such, the SGTs are trained with multiple passes over the training set---similar to how neural networks are trained for multiple epochs on the same training data.

\section{Experiments}
\label{sec:experiments}
This section demonstrates the efficacy of SGTs in the streaming classification, streaming regression, and batch multi-instance learning settings. We implemented the algorithm in Java, making use of the MOA framework~\citep{bifet2010a} for the experiments with incremental learning, and WEKA~\citep{hall2009} for the multi-instance learning experiments. The implementation is available online.\footnote{\url{https://github.com/henrygouk/stochastic-gradient-trees}}

Information about the streaming classification and regression datasets used in the experiments can be found in Table~\ref{tbl:incremental-datasets}.

\begin{table}[t]
	\centering \caption{Details of the classification and regression datasets used for evaluating incremental decision tree learners. No entry in the \# Classes column indicates that the dataset is associated with a regression task.}
	\label{tbl:incremental-datasets}
	\begin{tabular}{lrrr}
		\toprule 
		Dataset & \# Features & \# Instances & \# Classes \\
		\midrule
		Higgs & 27 & 11,000,000 & 2 \\
		HEPMASS & 27 & 10,500,000 & 2 \\
		KDD'99 & 41 & 4,898,430 & 40 \\
		Covertype & 55 & 581,012 & 7 \\
		AWS Prices & 7 & 27,410,309 & - \\
		Airline & 13 & 5,810,462 & - \\
		Zurich & 13 & 5,465,575 & - \\
		MSD Year & 90 & 515,345 & - \\
		\bottomrule
	\end{tabular}
\end{table}

\subsection{Classification}
\begin{figure}[h!]
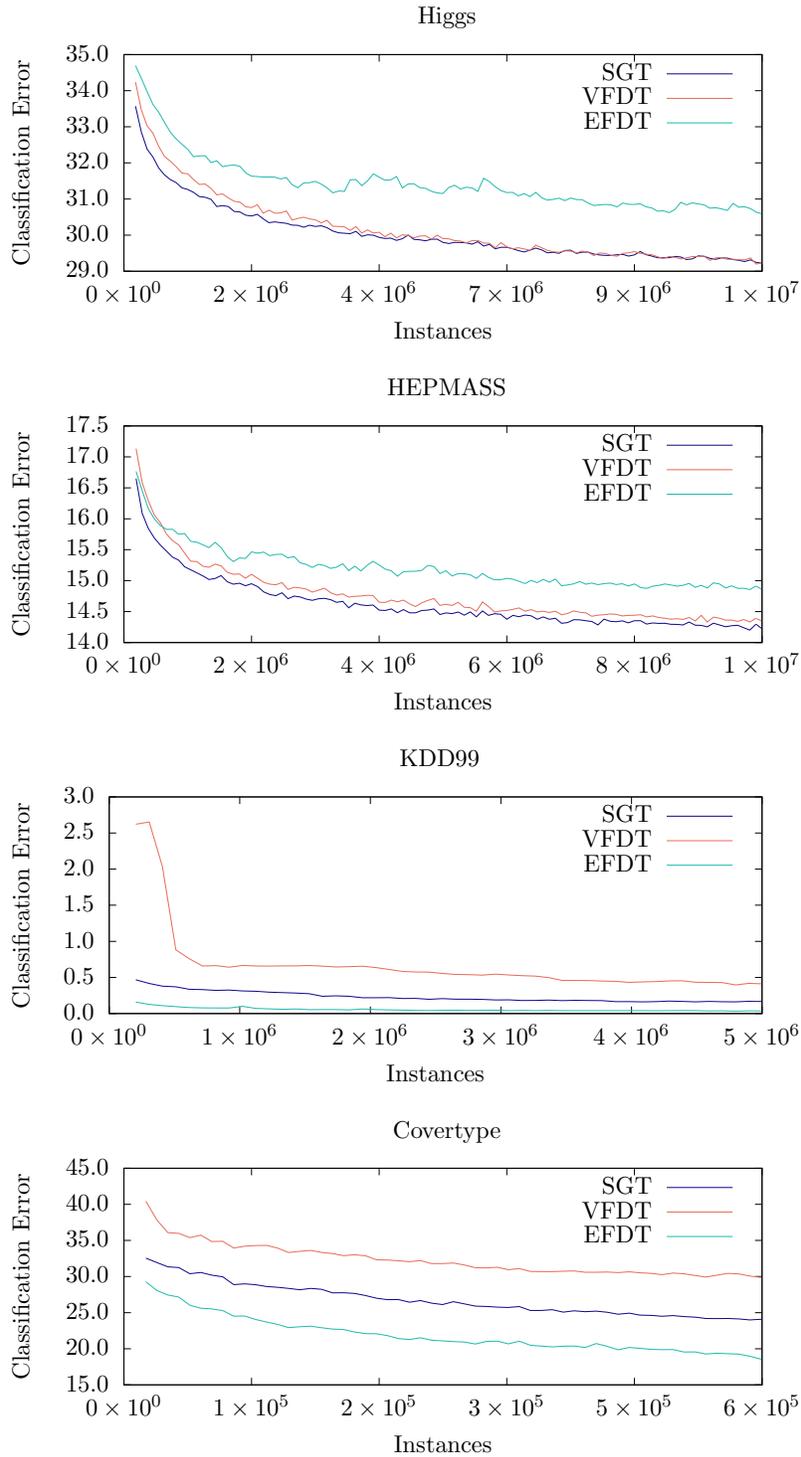

	\centering
	\resizebox{0.7\columnwidth}{!}{\input{higgs.tex}}
	\resizebox{0.7\columnwidth}{!}{\input{hepmass.tex}}
	\resizebox{0.7\columnwidth}{!}{\input{kdd99.tex}}
	\resizebox{0.7\columnwidth}{!}{\input{covertype.tex}}
	\caption{Learning curves for the incremental classification problems.}
	\label{fig:learning-curves-classification}
\end{figure}

For each dataset we report the mean classification error rate, model size, and runtime across 10 runs, where the data is randomly shuffled for each run. The standard Hoeffding tree algorithm (VFDT) and the more sample efficient extension of~\citet{manapragada2018} (EFDT) are used as baselines. Learning curves for these experiments are given in Figure~\ref{fig:learning-curves-classification}, and the numeric performance measurements are provided in Table~\ref{tbl:stream-results-class}. SGTs perform similarly to state of the art methods on classification problems, uniformly outperforming VFDT and exhibiting two wins and two losses each compared to EFDT. They also result is comparatively compact models, and are faster to train in the case where there is not a large number of classes.

\begin{table}[t]
	\centering \caption{Mean classification error, model size (number of nodes), and runtime (seconds) of the trees produced by the classification methods on 10 random shuffles of each dataset.}
	\label{tbl:stream-results-class}
	\begin{tabular}{crrrrr}
		\\
		\toprule
		 &  & Higgs & HEPMASS & KDD'99 & Covertype \\
		\midrule
		\multirow{3}{*}{Classification Error} & SGT & 30.10 & 14.66 & 0.27 & 26.94 \\
		 & VFDT & 30.25 & 14.81 & 0.73 & 32.54 \\
		 & EFDT & 31.46 & 15.22 & 0.07 & 22.05 \\
		 \midrule
		\multirow{3}{*}{Model size} & SGT & 1,933.2 & 1,289.0 & 446.3 & 620.8 \\
		 & VFDT & 8,081.4 & 7,256.0 & 160.0 & 91.8 \\
		 & EFDT & 38,535.1 & 24,760.2 & 913.8 & 3,261.4 \\
		\midrule
		\multirow{3}{*}{Runtime} & SGT & 384.13 & 290.46 & 466.26 & 32.56 \\
		 & VFDT & 115.68 & 108.11 & 33.15 & 3.04 \\
		 & EFDT & 512.32 & 425.49 & 97.61 & 47.23 \\
		 \bottomrule
	\end{tabular}
\end{table}

% Classification

\subsection{Regression}
\begin{figure}[h!]
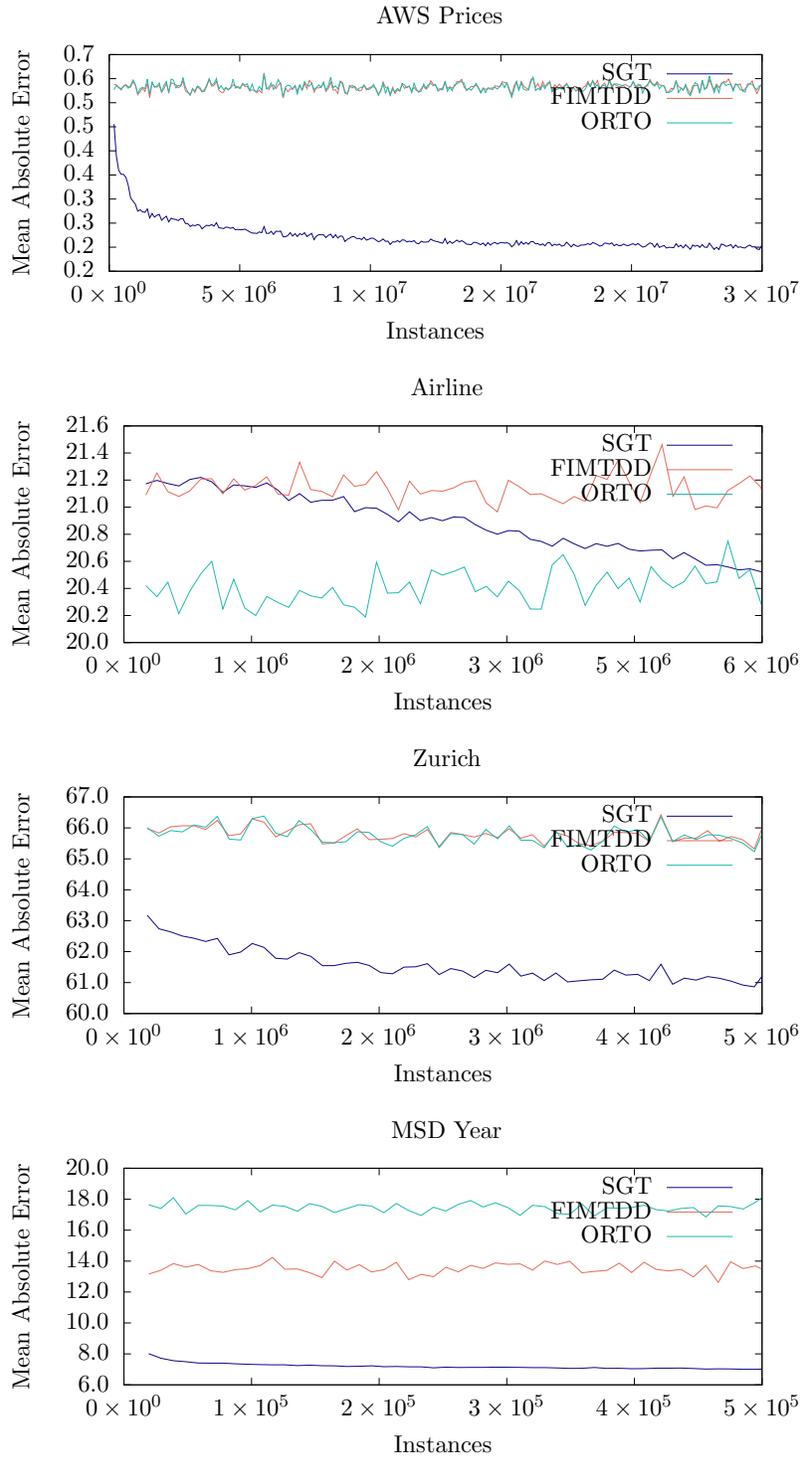

	\centering
	\resizebox{0.7\columnwidth}{!}{\input{aws.tex}}
	\resizebox{0.7\columnwidth}{!}{\input{airline.tex}}
	\resizebox{0.7\columnwidth}{!}{\input{zurich.tex}}
	\resizebox{0.7\columnwidth}{!}{\input{year-msd.tex}}
	\caption{Learning curves for the incremental regression problems.}
	\label{fig:learning-curves-regression}
\end{figure}

For the streaming regression experiments we report the mean absolute error, as well as the model size and runtime. As with the classification experiments, all measurements are averaged over 10 runs of each algorithm, and on each run the data is randomly shuffled. The FIMT-DD~\citep{ikonomovska2011a} and ORTO~\citep{ikonomovska2011} methods are used as points of reference for how state of the art streaming regression algorithms perform on the datasets considered. The learning curves for these experiments are given in Figure~\ref{fig:learning-curves-regression}, and the performance measurements are in Table~\ref{tbl:stream-results-regress}. These results show that, from a predictive performance point of view, SGTs generally outperform both FIMT-DD and ORTO. With the exception of the airline dataset, the final model sizes are drastically smaller than those of the baselines, and training time is comparable to FIMT-DD. Qualitatively, Figure~\ref{fig:learning-curves-regression} suggests that SGTs exhibit superior convergence properties, with the loss of the other methods plateauing very early on in training.

\begin{table}[t]
	\centering \caption{Mean absolute error, model size (number of nodes), and runtime (seconds) of the trees produced by the regression methods on 10 random shuffles of each dataset.}
	\label{tbl:stream-results-regress}
	\begin{tabular}{crrrrr}
		\toprule
		& & Airline & AWS Prices & Zurich & MSD Year \\
		\midrule
		\multirow{3}{*}{Mean Abs. Error} & SGT & 20.86 & 0.27 & 61.59 & 7.20 \\
		 & FIMT-DD & 21.14 & 0.58 & 65.80 & 13.52 \\
		 & ORTO & 20.41 & 0.58 & 65.78 & 21.76 \\
		 \midrule
		\multirow{3}{*}{Model size} & SGT & 39,287.0 & 3,316.3 & 763.1 & 235.8 \\
		 & FIMT-DD & 30,060.4 & 165,660.6 & 21,756.2 & 2,264.4 \\
		 & ORTO & 33,584.8 & 177,929.0 & 23,769.0 & 2,466.8 \\
		 \midrule
		\multirow{3}{*}{Runtime} & SGT & 22.76 & 79.96 & 33.18 & 58.23 \\
		 & FIMT-DD & 22.43 & 80.89 & 29.71 & 70.75 \\
		 & ORTO & 50.16 & 42.66 & 22.16 & 37.17 \\
		 \bottomrule
		\end{tabular}
\end{table}

\subsection{Multi-Instance Learning}
The evaluation metric used for multi-instance learning is the 10-fold cross-validation accuracy. SGTs are compared with two batch techniques specifically designed for MIL: the Quick Diverse Density Iterative (QDDI) approach of~\citet{foulds2010}, and an extension proposed by \citet{bjerring2011} to the multi-instance tree inducer (MITI) technique originally developed by~\citet{blockeel2005}. The results are reported in Table~\ref{tbl:multi-instance}, along with the average rank of each method. The multi-instance learning instantiation of our general purpose tree induction algorithm performs comparably with MITI---a state of the art tree induction technique for multi-instance learning---as shown by both the results of the hypothesis tests and the very similar average ranks. Both tree-based methods perform better than QDDI, as measured by average rank, and the hypothesis tests between QDDI and SGT indicate that SGT is more often the superior method.

\begin{table}[t]
\caption{\label{tbl:multi-instance}Accuracy of SGT, QDDI, and MITI on a collection of multi-instance classification datasets measured using 10-fold cross-validation. Statistically significant improvements or degredations in performance relative to SGT are denoted by $\circ$ and $\bullet$, respectively. The rank of each method on each dataset is given in parentheses.}
{\centering \begin{tabular}{lcc@{\hspace{0.1cm}}cc@{\hspace{0.1cm}}c}
\toprule
Dataset & SGT & QDDI & & MITI & \\
\midrule
atoms       & 73.36 (2) & 69.18 (3) &           & 84.06 (1) &   $\circ$\\
bonds       & 75.58 (2) & 72.43 (3) &           & 81.93 (1) &          \\
chains      & 81.99 (2) & 78.22 (3) &           & 88.42 (1) &          \\
component   & 91.92 (1) & 88.08 (3) & $\bullet$ & 89.49 (2) & $\bullet$\\
elephant    & 77.00 (2) & 81.00 (1) &           & 76.50 (3) &          \\
fox         & 55.00 (3) & 58.50 (2) &           & 60.00 (1) &          \\
function    & 95.75 (1) & 92.35 (3) & $\bullet$ & 94.85 (2) & \\
musk1       & 82.22 (2) & 77.00 (3) &           & 82.56 (1) & \\
musk2       & 71.64 (3) & 75.64 (1) &           & 74.64 (2) &          \\
process     & 96.42 (1) & 94.19 (3) & $\bullet$ & 96.17 (2) &          \\
tiger       & 77.00 (1) & 69.00 (3) &           & 76.00 (2) &          \\
thioredoxin & 87.08 (2) & 88.58 (1) &           & 80.87 (3) & \\
\midrule
Avg. Rank   & 1.83      & 2.42      &           & 1.75      & \\
\bottomrule
\end{tabular}
\footnotesize \par}
\end{table}

\section{Conclusion}
This paper presents the stochastic gradient tree algorithm for incrementally constructing a decision tree using stochastic gradient information as the source of supervision. In addition to showing how gradients can be used for building a single decision tree, we show how the Hoeffding inequality-based splitting heuristic found in many incremental tree learning algorithms can be replaced with a procedure based on the $t$-test, removing the requirement that one can bound the range of the metric used for measure the quality of candidate splits. Our experimental results on several different tasks demonstrate the generality of our approach, while also maintaining scalability and state of the art predictive performance. We anticipate that the algorithm presented in this paper will enable decision trees to tackle a diverse range of problems in future.

\acks{This research was supported by the Marsden Fund Council from Government funding, administered by the Royal Society of New Zealand.}

\bibliography{gouk395}

\end{document}